\pdfoutput=1

\documentclass[11pt]{article}

\usepackage[]{acl}

\usepackage{times}
\usepackage{latexsym}

\usepackage[T1]{fontenc}

\usepackage[utf8]{inputenc}

\usepackage{microtype}
\usepackage{booktabs}
\usepackage{graphics}
\usepackage{graphicx}
\usepackage{arydshln}
\usepackage{hyperref}
\usepackage{multirow}
\usepackage{footnote}
\usepackage{footmisc}
\usepackage{xcolor}
\usepackage{colortbl}
\usepackage{tabulary}
\usepackage{amsmath}
\usepackage{tablefootnote}

\title{How Does In-Context Learning Help Prompt Tuning?}

\author{
  Simeng Sun$^1$\hspace{3mm} Yang Liu$^2$\hspace{3mm} Dan Iter$^2$\hspace{3mm} Chenguang Zhu$^2$\hspace{3mm} Mohit Iyyer$^1$\\
  University of Massachusetts Amherst$^1$ \hspace{1em} Microsoft Research$^2$ \hspace{1em} \\
  \texttt{\{simengsun, miyyer\}@umass.edu} \\
  \texttt{\{yaliu10,iterdan,chezhu\}@microsoft.com} \\
  }

\begin{document}
\maketitle

\begin{abstract}
Fine-tuning large language models is becoming ever more impractical due to their rapidly-growing scale. This motivates the use of parameter-efficient adaptation methods such as prompt tuning (PT), which adds a small number of tunable embeddings to an otherwise frozen model, and in-context learning (ICL), in which demonstrations of the task are provided to the model in natural language without any additional training. Recently,~\citet{llm-clinical} propose ``instruction prompt tuning'' (IPT), which combines PT with ICL by concatenating a natural language demonstration with learned prompt embeddings. While all of these methods have proven effective on different tasks, how they interact with each other remains unexplored. In this paper, we empirically study when and how in-context examples improve prompt tuning by measuring the effectiveness of ICL, PT, and IPT on five text generation tasks with multiple base language models. We observe that 
(1) IPT does \emph{not} always outperform PT, and in fact requires the in-context demonstration to be semantically similar to the test input to yield improvements;  
(2) PT is unstable and exhibits high variance, but combining PT and ICL (into IPT) consistently reduces variance across all five tasks; and
(3) prompts learned for a specific source task via PT exhibit positive transfer when paired with in-context examples of a different target task.
Our results offer actionable insights on choosing a suitable parameter-efficient adaptation method for a given task.
\end{abstract}
\section{Introduction}

As large language models (LLMs) continue to grow in scale \citep{gpt3,chowdhery2022palm}, it is quickly becoming infeasible to fine-tune all of their parameters to solve a new task. As such, developing methods that \emph{efficiently} adapt LLMs to downstream tasks is critical. In this paper, we study the relationship between three such methods:

\begin{figure}
    \centering
    \includegraphics[width=\linewidth]{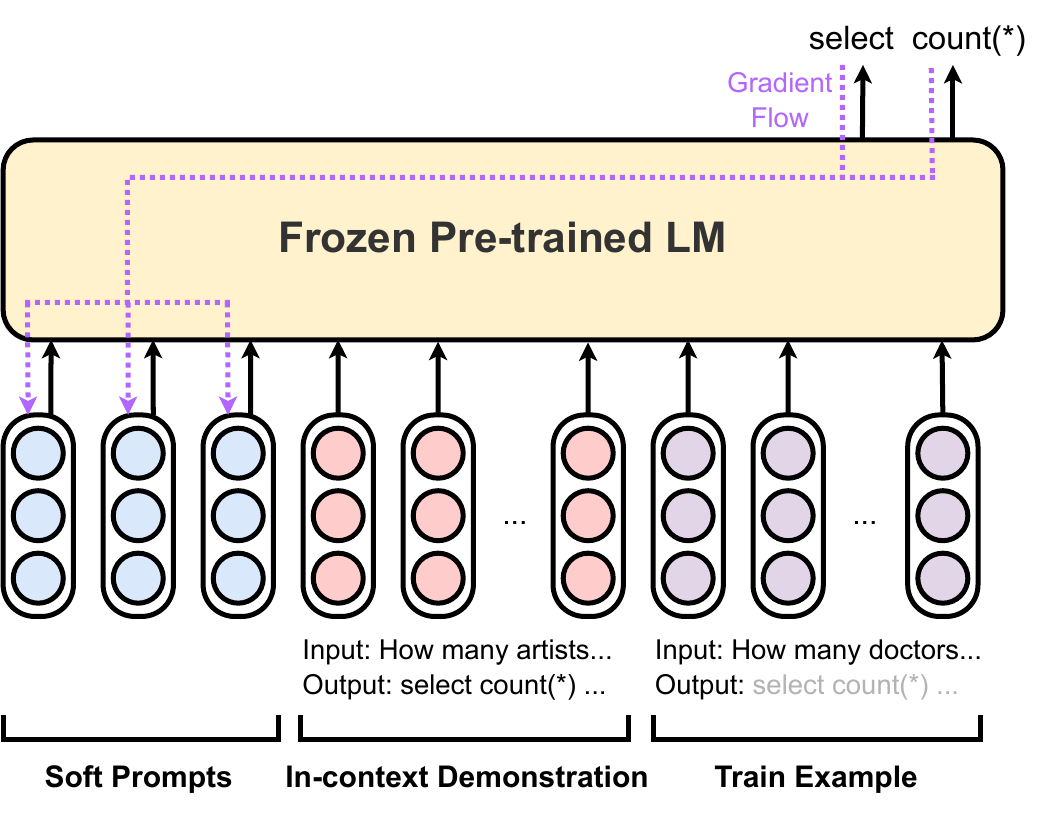}
    \caption{An illustration of instruction prompt tuning (IPT). Soft tunable prompt embeddings are prepended to a retrieved in-context demonstration, which is followed by the training example. In this paper, we study the mutual effect of the soft prompts and the discrete demonstrations in instruction prompt tuning. }
    \label{fig:fig_1}
\end{figure}

\begin{itemize}
    \item \textbf{In-context learning (ICL)}: The simplest method is to leverage \emph{in-context learning}, in which LLMs are prompted with instructions or demonstrations to solve a new task without any additional training~\citep{gpt3}. ICL can be further improved by dynamically retrieving demonstrations that are similar to a particular test input, rather than choosing demonstrations at random~\citep{good-gpt3-example}. However, it still struggles on complex and out-of-domain downstream tasks~\citep{input-tuning,liu2022fewshot}.
    \item \textbf{Prompt tuning (PT)}: The limitations of ICL beg the question of whether a small amount of training can help. In \emph{prompt tuning}, the vast majority of the LLM is kept frozen while a small number of new tunable embeddings are concatenated to every test input~\cite{lester-pt}. While PT generally outperforms ICL, it is unstable and difficult to optimize~\citep{peft-study}.
    \item \textbf{Instruction prompt tuning (IPT)}: More recently, \citet{llm-clinical} combine ICL and PT into \emph{instruction prompt tuning}, which concatenates retrieved in-context demonstrations with tunable prompt embeddings, and they show its effectiveness at adapting LLMs to the medical domain. 
\end{itemize}


Little is known about the conditions in which any of these methods outperforms the other; more generally, the mutual effect of in-context learning and prompt tuning remains understudied. We shed light on these questions by comparing ICL, PT, and IPT across five text generation tasks using three base LMs of comparable size (BLOOM 1.1B, OPT 1.3B, and GPT2 XL 1.5B). We focus mainly on out-of-distribution tasks that challenge the limits of parameter-efficient adaptation methods, including ToTTo~\citep{totto} and DART~\citep{dart} for data-to-text generation, Logic2Text~\citep{logic2text} for logic-to-text generation, and Spider~\citep{spider} and MTOP~\citep{mtop} for semantic parsing.

We summarize our findings as follows:
\begin{itemize}
    \item Both PT and IPT consistently outperform ICL across all five tasks. This result demonstrates the value of training at least a small set of parameters for out-of-domain tasks. 
    \item That said, there is no clear winner between PT and IPT, as performance is highly dependent on the task and experimental configuration (e.g., number of tunable embeddings).
    \item IPT outperforms PT on examples for which the in-context demonstration is highly similar to the test input. The most striking case of this is ToTTo, where IPT is significantly better than PT; we attribute this result to overlapping train/test tables in the dataset as well as formulaic output.
    \item PT exhibits high variance, especially when there are more tunable parameters. IPT reduces variance, and its performance is less dependent on the number of prompt embeddings than PT.
    \item While prompt embeddings learned via PT cannot be directly transferred to unseen tasks, we discover that they are transferable to new tasks given in-context demonstrations, and that combining source task prompts with target task demonstrations outperforms ICL in this transfer setting.
\end{itemize}

\section{Background}

Parameter-efficient fine-tuning methods~\citep{pmlr-v97-houlsby19a,karimi-mahabadi-etal-2021-parameter,bitfit} specialize LLMs to a target task while keeping most of their parameters frozen and adjusting just a small number of task-specific parameters. Since full-model fine-tuning is prohibitively expensive on consumer-grade hardware for most LLMs, such methods increase the accessibility of LLM research and deployment. Here, we give a more formal specification of the parameter-efficient tuning methods that we experiment with in this paper.

\paragraph{In-context learning:} ~\citet{gpt3} show that their 175B-parameter GPT-3 model is capable of solving unseen tasks by leveraging information from in-context instructions (\emph{zero-shot}) and/or demonstrations (\emph{few-shot}). Inserting $k$ in-context input-output pairs $[\mathbf{X}_{icl};~\mathbf{Y}_{icl}]$ before the test input significantly improves the performance of solving a target task:
\begin{align*}
    \text{Input}_{\text{ICL}} = \text{concat}\big( [\mathbf{X}_{icl};~\mathbf{Y}_{icl}]_{1}^{k};~\mathbf{X}_{test}\big)
\end{align*}
Recent studies propose approaches that discover better in-context demonstrations by retrieving examples semantically similar to each test input~\citep{good-gpt3-example}, as well as eliciting chain-of-thought reasoning~\citep{wei2022chain} and breaking tasks into sub-problems with least-to-most prompting~\citep{least-to-most}.

\paragraph{Prompt tuning:} In-context learning struggles on out-of-domain tasks, which motivates alternate approaches that tune a small fraction of the LLM's parameters~\citep{peft-study}. In this paper, we focus on prompt tuning~\cite{lester-pt,p-tuning}, which prepends soft tunable prompt embeddings to the input tokens $\mathbf{X}_{test}$. Since it only modifies the input to the LLM, it is easy to implement and, unlike adapter-based approaches~\citep{adapter-nmt}, does not change the internal model structure. Formally, let $\mathbf{E} = \{\mathbf{e_1}, \dots, \mathbf{e_k}\}$ be a sequence of new tunable prompt embeddings, while $\mathbf{X} = \{\mathbf{x_1},\dots,\mathbf{x_m}\}$ and $\mathbf{Y} = \{\mathbf{y_1},\dots,\mathbf{y_n}\}$ denote the token embeddings of the input and output of an example, respectively. Then, the input to prompt tuning at inference time can be expressed as
\[\text{Input}_{\text{PT}} = \text{concat}\big(\mathbf{E};~\mathbf{X}_{test}\big).\]

Since prompt tuning requires training the tunable embeddings $\mathbf{E}$, we require access to a training set $\mathbf{X}_{train}$ for the target task, unlike with in-context learning. While training $\mathbf{E}$, we feed $\mathbf{X}_{train}$ to the LLM as input, and the loss is computed over corresponding output tokens $\mathbf{Y}_{train}$ that follows $\mathbf{X}_{train}$. 

\paragraph{Instruction Prompt Tuning.} More recently, ~\citet{llm-clinical} proposes instruction prompt tuning, which concatenates the soft prompts with hard in-context demonstrations.
Using the notation from above, the input of IPT is:
\[\text{Input}_{\text{IPT}}= \text{concat}\big(\mathbf{E};~[\mathbf{X}_{icl};~\mathbf{Y}_{icl}]_{1}^{k};~\mathbf{X}_{test}\big).\]
Note that in our experiments, the prompt embeddings $\mathbf{E}$ are trained specifically for a single downstream task, whereas ~\citet{llm-clinical} share them across multiple tasks in the medical domain. Also, this kind of hybrid of soft and hard prompt tokens has been previously employed by~\citet{ppt} and~\citet{ptr}. Instruction prompt tuning resembles MetaICL~\citep{min-etal-2022-metaicl} and in-context tuning~\citep{chen-etal-2022-meta} in that in-context demonstrations are part of the input during training; however, IPT tunes just the prompt embeddings rather than the full model. 

\begin{table}[!t]
\centering
\footnotesize
\begin{tabular}{@{}lrrrr@{}}
\toprule
\multicolumn{1}{l}{} & \#\textbf{Train} & \#\textbf{Test} & \begin{tabular}[c]{@{}c@{}}\textbf{Avg. len} \\ X$_{PT}$\end{tabular} & \begin{tabular}[c]{@{}c@{}}\textbf{Avg. len} \\ X$_{IPT}$\end{tabular} \\ \midrule
ToTTo                & 120,761  & 7,700   & 95                                                      & 202                                                      \\
DART                 & 62,659   & 5,097   & 41                                                      & 106                                                      \\
Spider               & 7,000    & 1,034   & 109                                                     & 244                                                      \\
MTOP                 & 15,667   & 2,235   & 680                                                     & 1,390                                                     \\
Logic2Text           & 8,566    & 1,095   & 56                                                      & 136                                                      \\ \bottomrule
\end{tabular}
\caption{Dataset statistics. We provide the average length of each example for both prompt tuning and instruction prompt tuning. IPT has a longer input length on average because one retrieved demonstration is included  with the soft prompt and the test input.\tablefootnote{Due to the longer input length, we notice IPT takes longer to train than PT.}}
\label{tab:stats}
\end{table}

\begin{table*}
\centering
\scalebox{0.99}{\begin{tabular}{@{}lccccc@{}}
\toprule
 & \begin{tabular}[c]{@{}c@{}}ToTTo \\ (BLEU)\end{tabular} & \begin{tabular}[c]{@{}c@{}}Dart \\ (BLEU)\end{tabular} & \begin{tabular}[c]{@{}c@{}}Spider \\ (Exact Match)\end{tabular} & \begin{tabular}[c]{@{}c@{}}Mtop \\ (Exact Match)\end{tabular} & \begin{tabular}[c]{@{}c@{}}Logic2text \\ (BLEC)\end{tabular} \\ \midrule
 \textbf{BLOOM-1.1B} & & & & & \\
\hspace{1em} random one-shot ICL             & 5.8       & 8.3      & 0.4      & 0.0    & 37.6           \\
\hspace{1em} retrieved one-shot ICL           & 35.1      & 23.9     & 3.9      & 18.5   & 70.1           \\
\hspace{1em}  retrieve three-shot ICL & 41.3 & 29.7 & 5.0 & 12.7 & 71.0 \\
\midrule
\textbf{BLOOM-1.1B} & & & & & \\
\hspace{1em} Prompt Tuning & 36.3$_{\pm 0.3}$      & 41.2$_{\pm 0.9}$     & 35.5$_{\pm 1.6}$   & 25.2$_{\pm 16.4}$   & 87.6$_{\pm 1.5}$           \\
\hspace{1em} Instruction Prompt Tuning  & 47.1$_{\pm 0.2}$      & 41.4$_{\pm 0.1}$     & 33.2$_{\pm 1.1}$   &  62.6$_{\pm 0.7}$  & 86.4$_{\pm 1.1}$           \\\midrule
\textbf{OPT-1.3B} & &  & & & \\
\hspace{1em} Prompt Tuning  & 38.5$_{\pm 1.0}$ & 44.5$_{\pm 0.2}$ & 14.4$_{\pm 2.3}$ & 6.4$_{\pm 6.5}$ & 80.6$_{\pm 3.7}$  \\
\hspace{1em} Instruction Prompt Tuning  & 46.3$_{\pm 0.9}$ & 42.9$_{\pm 0.4}$ & 14.2$_{\pm 2.1}$& 10.4$_{\pm 6.5}$ &  84.6$_{\pm 1.0}$  \\\midrule
\textbf{GPT-2-XL-1.5B} & &  & & & \\
\hspace{1em} Prompt Tuning  & 37.3$_{\pm 0.2}$ & 43.5$_{\pm 0.2}$  & 27.0$_{\pm 2.1}$ & 41.4$_{\pm 5.6}$ & 87.2$_{\pm 1.6}$  \\
\hspace{1em} Instruction Prompt Tuning  & 48.0$_{\pm 0.0}$ & 42.1$_{\pm 0.2}$ & 23.0$_{\pm 0.1}$  & 19.8$_{\pm 14.9}$ & 85.8$_{\pm 1.5}$   \\
\bottomrule
\end{tabular}}
\caption{Providing a retrieved in-context demonstration significantly outperforms a random in-context training demonstration, although both PT and IPT generally outperform ICL. Here, we only report the performance of PT and IPT with 25 tunable prompt embeddings. Tuning the number of prompt embeddings further improves performance for both methods, as shown in Figure~\ref{fig:vpt}. }
\label{tab:bloom_summ}
\end{table*}

\begin{figure*}[!t]
    \centering
    \includegraphics[width=0.3\linewidth]{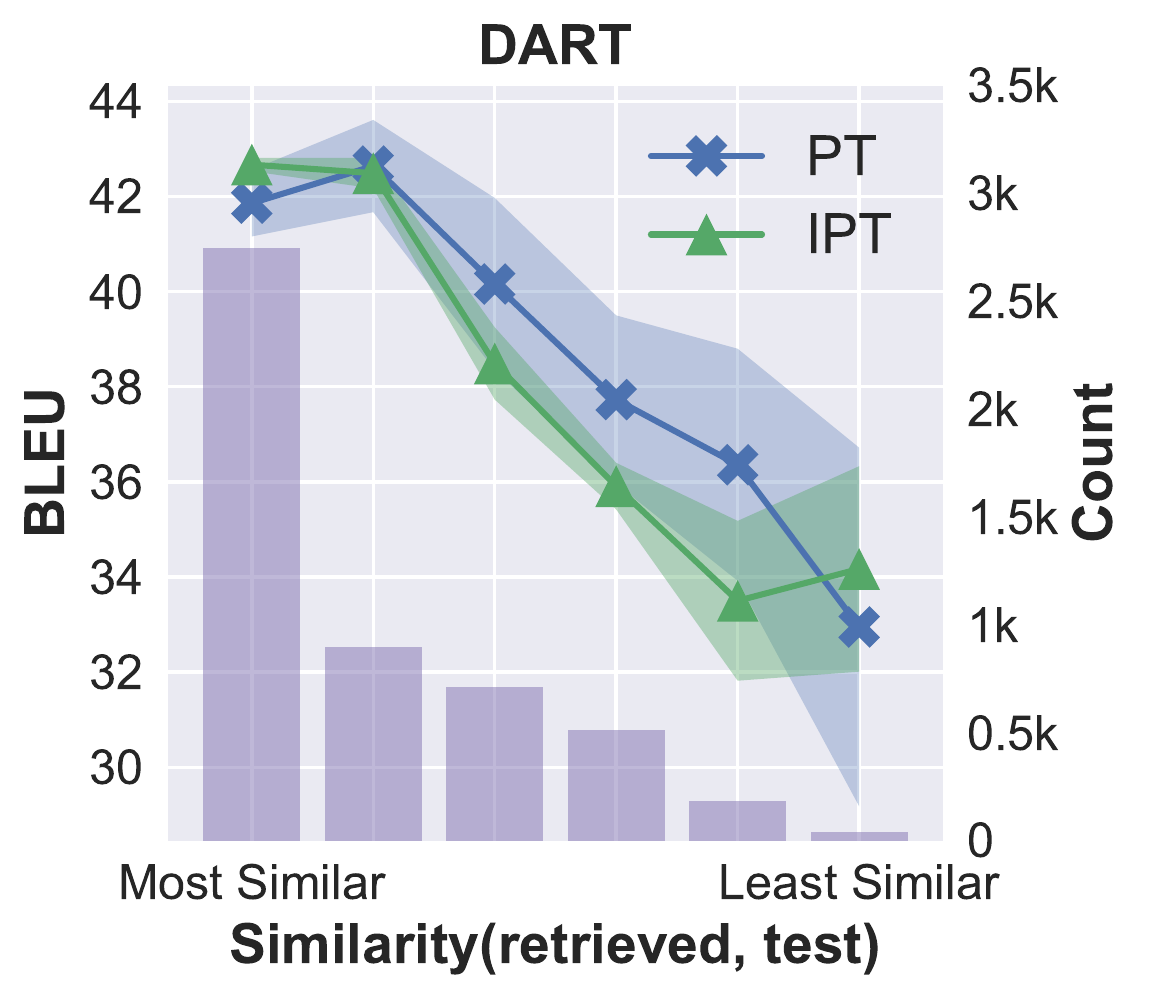}\includegraphics[width=0.3\linewidth]{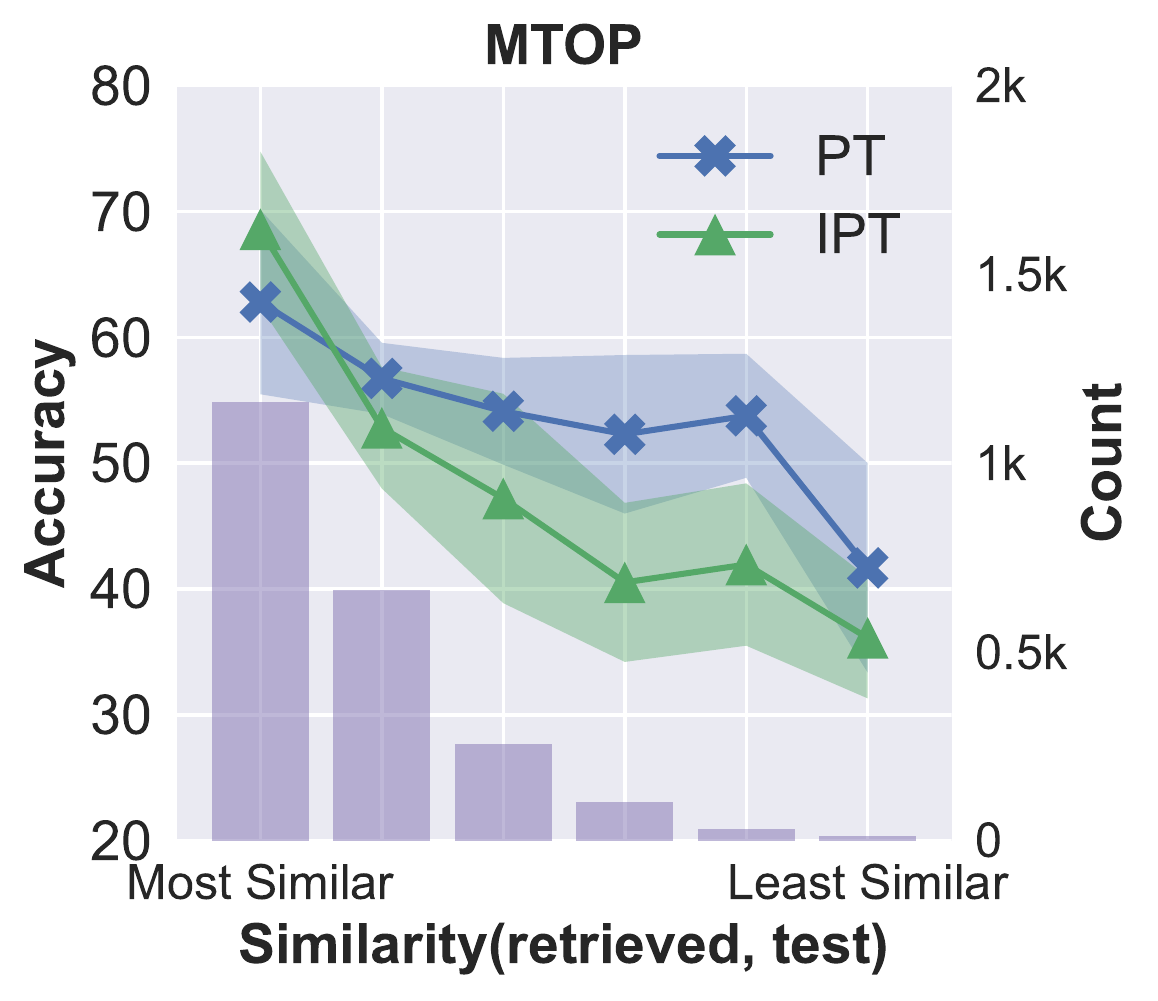}\includegraphics[width=0.3\linewidth]{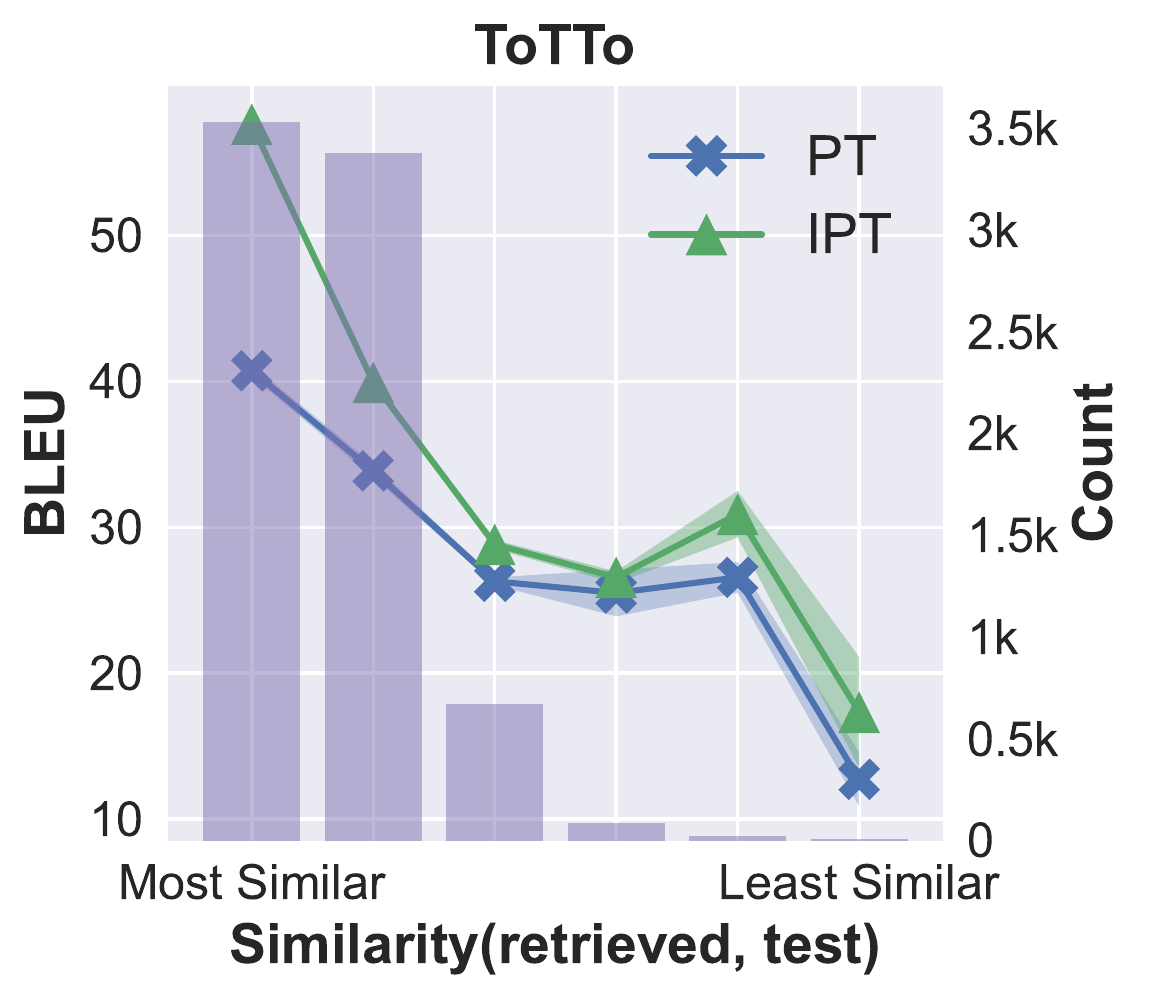}
    \caption{IPT performs better than PT on examples for which the input of retrieved in-context demonstration is very similar to the test input. However, IPT degrades quickly as the retrieved demonstration becomes less similar, and for both DART and MTOP it underperforms PT on out-of-distribution test inputs. Over 85\% of test inputs in ToTTo have highly-similar training examples, which is an explanation for IPT's significantly higher performance on ToTTo.}
    \label{fig:sim}
\end{figure*}

\begin{figure*}
    \centering
    \includegraphics[width=0.2\textwidth]{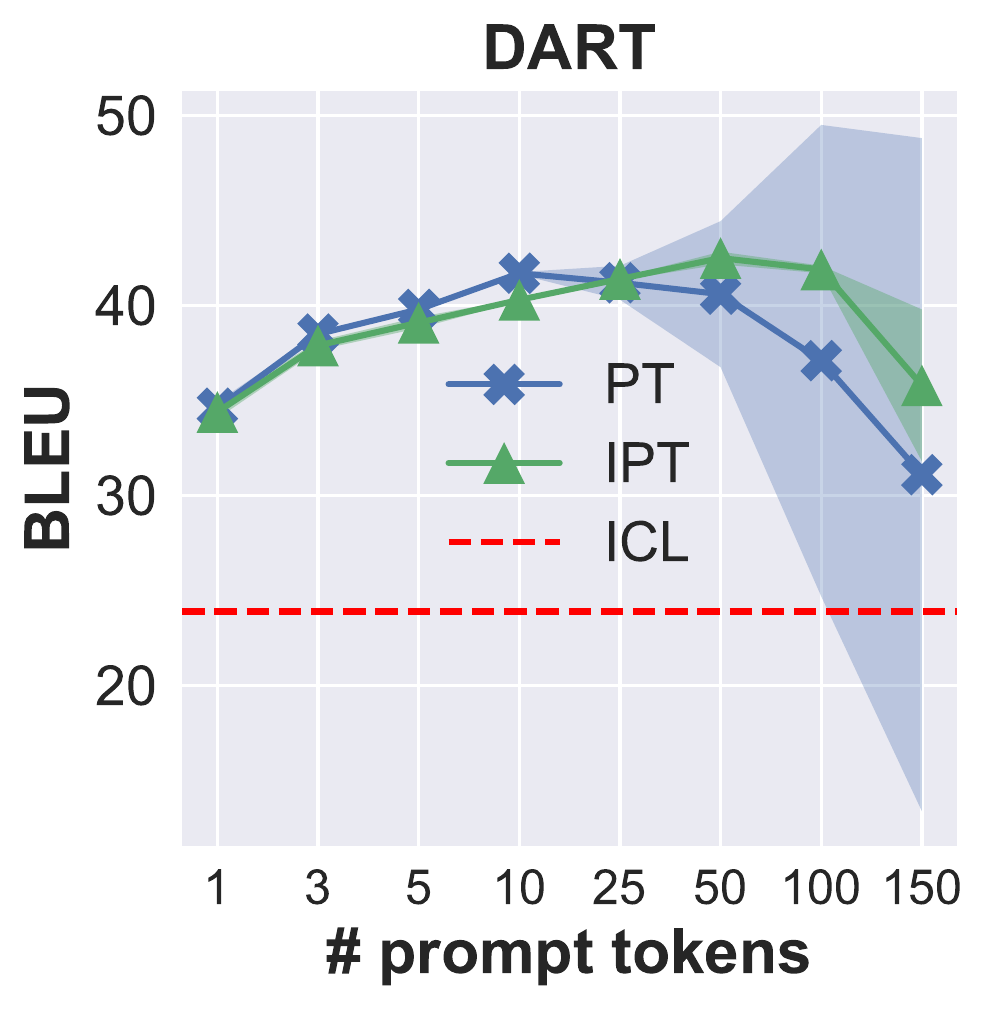}\includegraphics[width=0.2\textwidth]{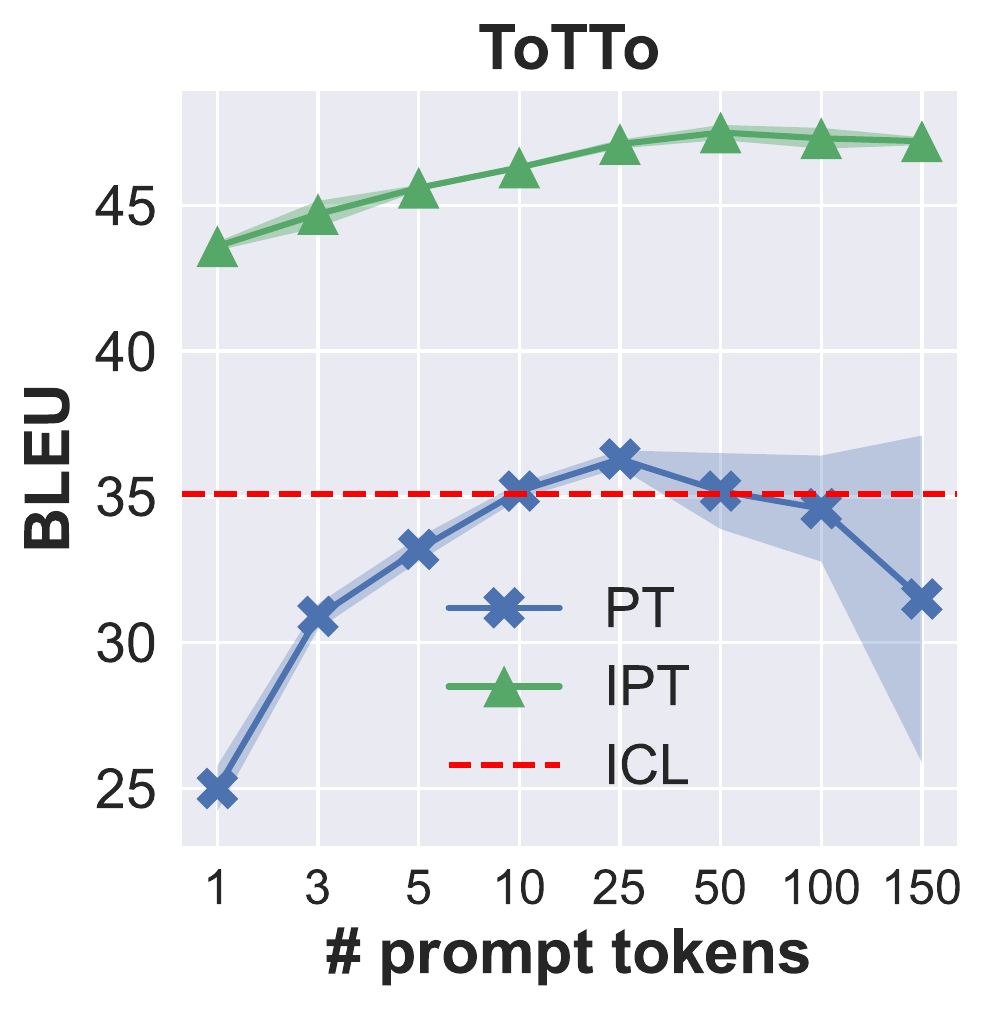}\includegraphics[width=0.2\textwidth]{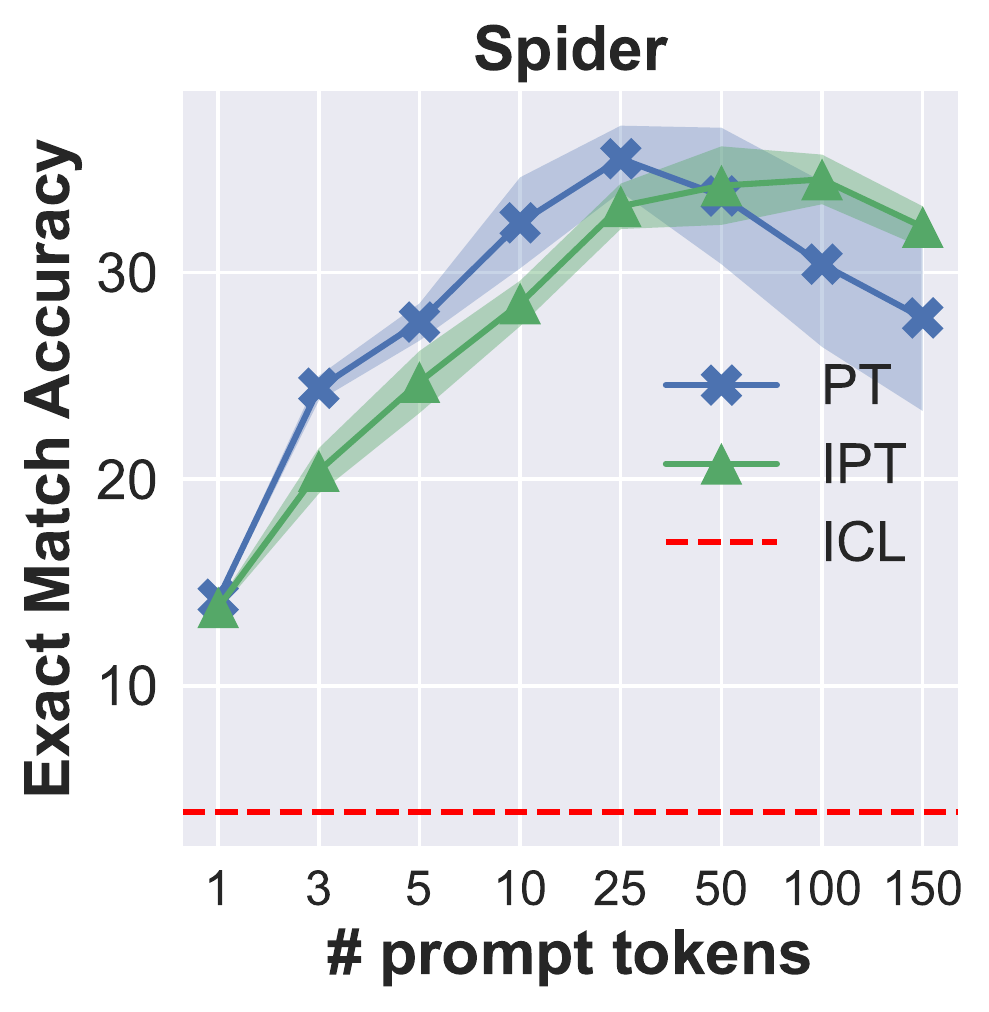}\includegraphics[width=0.2\textwidth]{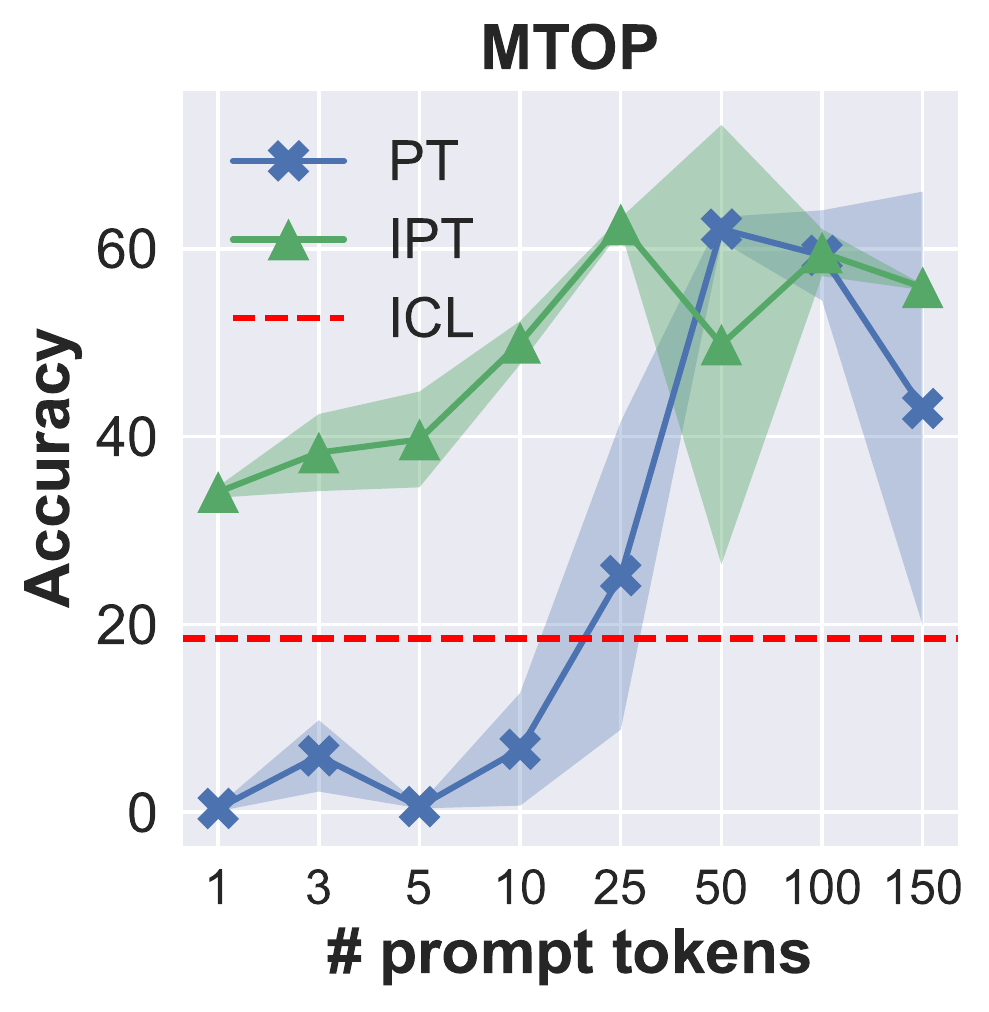}\includegraphics[width=0.2\textwidth]{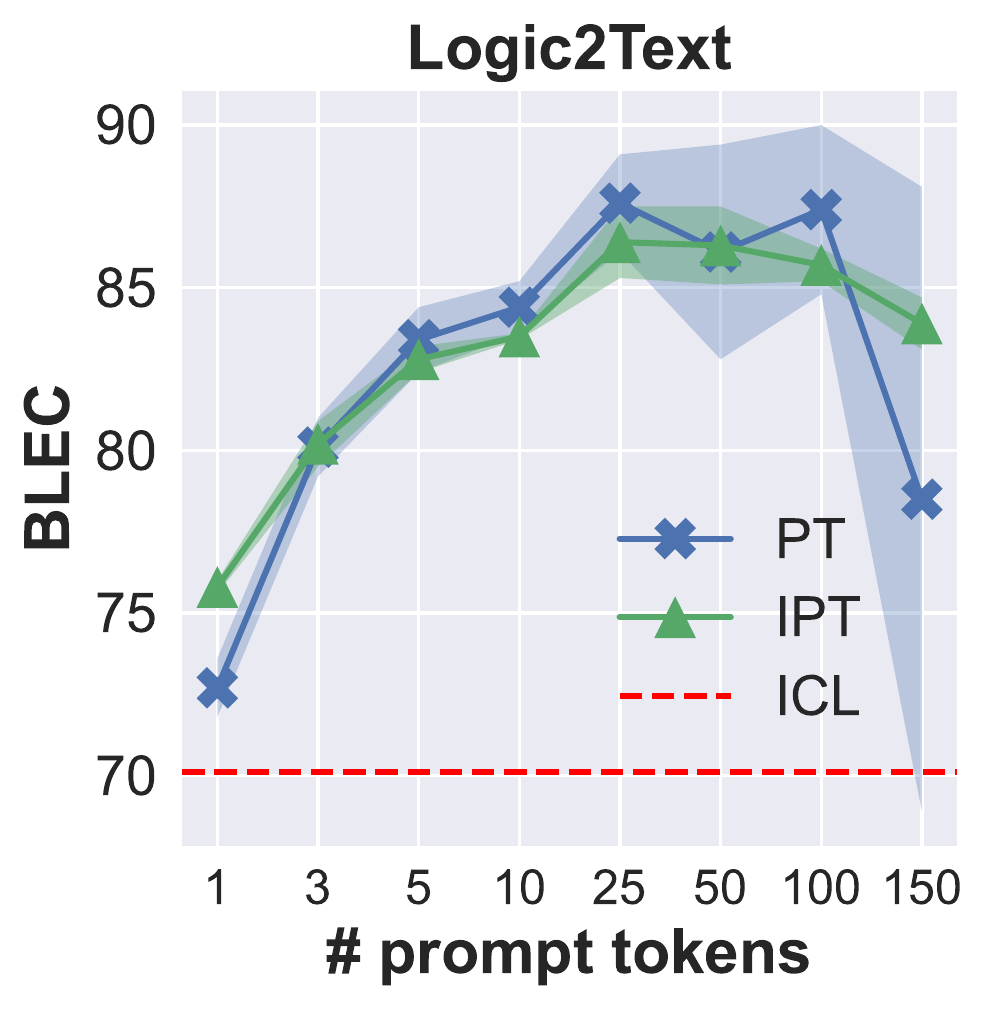}
    \caption{Comparing the performance of prompt tuning, instruction prompt tuning, and in-context learning, where the latter two methods are provided with one retrieved in-context demonstration, on five language generation tasks varying the number of soft prompt tokens. 
    The best PT and IPT configurations always outperform ICL. PT exhibits increasing variance as the number of tunable parameters increases, whereas IPT is relatively more stable. IPT is less sensitive overall to the number of prompt tokens, which makes it preferable in situations where hyperparameter tuning is computationally expensive.}
    \label{fig:vpt}
\end{figure*}

\section{Experimental setup} \label{sec:exp}
How can a soft prompt benefit from the added information provided by a retrieved in-context demonstration? To answer this question, we run experiments comparing the performance of ICL, PT, and IPT across a variety of different tasks, configurations, and base language models.
While past research into prompt tuning has mostly focused on natural language understanding tasks~\citep{lester-pt,tu-spot}, we focus on \emph{language generation} tasks in this paper, with a specific focus on tasks where either the input or output is (relatively) out-of-domain for the pretrained LLM. 

\paragraph{Datasets:} We explore three kinds of tasks: data-to-text generation, logic-to-text generation, and semantic parsing. In \emph{data-to-text} generation, the input is of structured data, either expressed as sets of triplets as in \textbf{DART}~\citep{dart} or as linearized table strings as in \textbf{ToTTo}~\citep{totto}. The output of both tasks are short sentences describing the data or table, which is evaluated with BLEU~\citep{bleu}. For \emph{semantic parsing}, in which a natural language utterance is mapped to a logical form, we evaluate on \textbf{Spider}~\citep{spider} and \textbf{MTOP}~\citep{mtop} and report exact match accuracy. Finally, in the \textbf{Logic2Text} \emph{logic-to-text} task~\citep{logic2text}, we use the metric BLEC to be consistent with other works~\cite{unifiedskg}.\footnote{For Spider, MTOP, and Logic2Text, we include knowledge information, such as linearized table schema, before the textual input. We use the processed data in~\href{https://github.com/HKUNLP/UnifiedSKG}{https://github.com/HKUNLP/UnifiedSKG}. For ToTTo, we use the processed data provided by~\citet{good-gpt3-example}.} More details about each dataset are presented in Table~\ref{tab:stats}.

\paragraph{Models:} We experiment with the BLOOM-1.1B\footnote{\url{https://huggingface.co/bigscience/bloom-1b1}}, OPT-1.3b\footnote{\url{https://huggingface.co/facebook/opt-1.3b}}, and GPT-2-XL-1.5B\footnote{\url{https://huggingface.co/gpt2-xl}} models on all our tasks. For our fine-grained analysis, we focus on the BLOOM checkpoint, which has 24 Transformer layers, an embedding dimensionality of 1536, and 16 attention heads, and is trained on multilingual text as well as programming language corpora.\footnote{\url{https://huggingface.co/spaces/bigscience/BigScienceCorpus}} For stabler and faster prompt tuning convergence, we employ the reparameterization trick introduced by~\citet{prefix-tune} by adding two feed-forward layers atop the initial prompt embeddings; the transformed prompt embeddings are then fed as input to the model.\footnote{Unlike~\citet{p-tuning-v2}, we modify only the input layer of the language model instead of every layer. A similar approach is also used by~\citet{input-tuning}.} For both PT and IPT, we randomly initialize all prompt embeddings, use a batch size of 8, and evaluate the best checkpoint selected by dev loss after training for 5 epochs with the AdamW optimizer~\citep{adamw}. The learning rate and weight decay for each task are provided in Appendix~\ref{sec:hparam}. For each configuration, we report the averaged performance over three runs. 

\paragraph{In-context demonstration retrieval.} 
Following~\citet{good-gpt3-example}, we use dense retrieval to select good in-context examples for instruction prompt tuning. We encode the input of each example with a large language model\footnote{We use GPT-neo-1.3b \url{https://huggingface.co/EleutherAI/gpt-neo-1.3B} in our experiment.} and extract the last token representation as the dense representation for the encoded sequence. We then use FAISS~\citep{johnson2019billion}\footnote{~\url{https://github.com/facebookresearch/faiss}} to retrieve the nearest-neighbor training example as an in-context demonstration.\footnote{To avoid the order of in-context examples~\citep{good-gpt3-example} complicating the experiments, we only provide one in-context demonstration per example. More details about retrieving examples for DART are included in Appendix~\ref{sec:appendix-dart}.}

\section{Analysis}
Table~\ref{tab:bloom_summ} shows that both PT and IPT (with 25 soft prompt tokens each) significantly outperform ICL with randomly retrieved in-context demonstration on all five tasks, which supports conclusions drawn from prior studies on prompt tuning. The performance of ICL can be further improved by having ``good'' retrieved in-context demonstrations, however it still lags behind PT and IPT on most tasks. On the other hand, there is no such clear trend in the relative performance of PT and IPT, other than on the ToTTo dataset, where IPT is a clear winner. We discover that the in-context demonstration included in IPT is helpful when the test input and the demonstration are semantically similar; if they are too different, then the demonstration can actually hurt. We also find that IPT consistently reduces variance across all tasks, indicating that the additional in-context example improves the stability of prompt tuning. Finally, we experiment with the \emph{transferability} of soft prompts trained on a source task and then used for a different target task. We observe improvements over ICL when concatenating an in-context demonstration of the target task with a soft prompt trained on a different source task.

\begin{table*}[]
    \centering
    \footnotesize
    \scalebox{0.9}{\begin{tabular}{@{}lll@{}}
\toprule
                   & \textbf{Input}                                  & \textbf{Output}  \\ \midrule
\textbf{Retrieved} & \begin{tabular}[c]{@{}l@{}}\textless{}page\_title\textgreater List of Governors of South Carolina\\   \textless{}section\_title\textgreater Governors under the Constitution of 1868  \\ \textless{}table\textgreater \textless{}cell\textgreater 80 \textless{}col\_header\textgreater \#  \textless{}col\_header\textgreater 74  \textless{}col\_header\textgreater 75  \\ \textless{}col\_header\textgreater 76  \textless{}col\_header\textgreater 77  \textless{}col\_header\textgreater 78  \textless{}col\_header\textgreater 79   \\ \textless{}cell\textgreater Johnson Hagood \textless{}col\_header\textgreater Governor  \\ \textless{}row\_header\textgreater 80 \textless{}/row\_header\textgreater  \textless{}cell\textgreater November 30, 1880 \\ \textless{}col\_header\textgreater Took Office  \textless{}row\_header\textgreater 80 \textless{}/row\_header\textgreater  \\ \textless{}cell\textgreater December 1, 1882 \textless{}col\_header\textgreater Left Office  \textless{}row\_header\textgreater 80 \textless{}/row\_header\textgreater{}\end{tabular} & \begin{tabular}[c]{@{}l@{}}Johnson Hagood was \\ the 80th Governor of \\ South Carolina from 1880 to 1882.\end{tabular}   \\ \midrule
\textbf{Test}      & \begin{tabular}[c]{@{}l@{}}\textless{}page\_title\textgreater List of Governors of South Carolina\\   \textless{}section\_title\textgreater Governors under the Constitution of 1868  \\ \textless{}table\textgreater \textless{}cell\textgreater 76 \textless{}col\_header\textgreater \#  \textless{}col\_header\textgreater 74  \\ \textless{}col\_header\textgreater 75   \textless{}cell\textgreater Daniel Henry Chamberlain \\ \textless{}col\_header\textgreater Governor  \textless{}row\_header\textgreater 76 \textless{}/row\_header\textgreater  \\ \textless{}cell\textgreater December 1, 1874 \textless{}col\_header\textgreater Took Office  \\ \textless{}row\_header\textgreater 76 \textless{}/row\_header\textgreater{}\end{tabular}                & \begin{tabular}[c]{@{}l@{}}Daniel Henry Chamberlain was \\ the 76th Governor of \\ South Carolina from 1874.\end{tabular} \\ \bottomrule
\end{tabular}}
    \caption{An example from ToTTo dev set and its corresponding top retrieved in-context example. IPT and in-context learning have a significant advantage over PT due to the presence of the in-context demonstration, which has high word overlap and follows the same template as the test output.}
    \label{tab:totto_eg}
\end{table*}

\paragraph{In-context learning underperforms prompt tuning:}
In line with experiments from prior work~\citep{liu2022fewshot}, we observe that ICL performs consistently worse than PT and IPT, even when using retrieved demonstrations instead of random demonstrations. This result shows the value of training a small number of new parameters to specialize a language model to the target task, especially for out-of-distribution generation. The lone exception is ToTTo, for which ICL is competitive with PT; we discuss reasons for this and the improvements from IPT later in this section. 

\paragraph{No clear winner between PT and IPT:}
Despite receiving additional signal from the retrieved in-context demonstration, IPT does not consistently outperform PT. Our results in Table~\ref{tab:bloom_summ}, also visualized for the BLOOM-1.1B model in Figure~\ref{fig:vpt}, show that IPT is in fact worse on several task and model configurations. The relative performance of these two methods highly depends on the task and the number of tunable parameters.
For instance, IPT performs better than PT with OPT-1.3B on Logic2Text (84.6 vs. 80.6), whereas it is worse than PT if use GPT-2-XL as the base model (85.8 vs. 87.2). 

\paragraph{IPT helps when the in-context demonstration is similar to the test input:} \label{sec:analysis-sim}
Clearly in-context demonstrations can work synergistically with soft prompts in some cases (e.g., on ToTTo), so under what conditions does this happen?
To understand the effect of in-context demonstrations in IPT, we divide each test set into multiple bins based on the \emph{semantic similarity} between the input of in-context example and the input of test example, and we evaluate model performance on each bin. 
More specifically, we encode the input of each example with large pre-trained LM by extracting the last token representation, and measure the similarity in latent space, which is also used for ICL demonstration retrieval as described in section~\ref{sec:exp}.
As shown for three different datasets in Figure~\ref{fig:sim}, the performance\footnote{
In Figure~\ref{fig:sim}, we select two task and model configurations on which IPT and PT achieve almost identical average performance (DART with 25 prompt tokens, and MTOP with 100 prompt tokens) while having the same number of tunable parameters.} of both PT and IPT decreases as the similarity of the retrieved in-context example input to the test input decreases. IPT outperforms PT when it is possible to retrieve highly-similar in-context examples (left-most bin of each plot). However, the performance of IPT degrades considerably as the in-context example becomes less similar, and PT outperforms IPT on both DART and MTOP on the most out-of-distribution examples (right-most bin). These results suggest that low-quality in-context examples can confuse the base LM, which motivates future work on dynamic methods that choose whether or not to include an in-context example based on thresholded similarity to the test input.

\paragraph{Overlap in ToTTo inflates IPT performance:} \label{sec:totto_analysis}
IPT significantly outperforms PT on ToTTo (e.g., 48.0 vs. 37.3 with GPT-2-XL) as shown in both Table~\ref{tab:bloom_summ} and Figure~\ref{fig:vpt}. We attribute this gap to substantial overlap between training and testing tables, along with very formulaic outputs. Table~\ref{tab:totto_eg} contains an example where the train and test input belong to the same parent page, and the output format is identical; all that is needed is to copy the training output and edit the named entities and numerics according to the table. This gives IPT a big advantage: as shown in the right-most plot of Figure~\ref{fig:sim}, IPT outperforms PT when the in-context demonstration is very similar to the evaluated input, which constitutes over 85\% of total evaluation examples in ToTTo.  On the other hand, when the in-context examples become less similar to the test input, PT and IPT achieve similar performance. As the large improvement on ToTTo comes mostly from these ``easy'' examples, we encourage future research in this domain to also evaluate on ``harder'' subsets where there is no table overlap, and also consider other more complex datasets, such as Spider and DART.

\begin{figure}
    \centering
    \includegraphics[width=\linewidth]{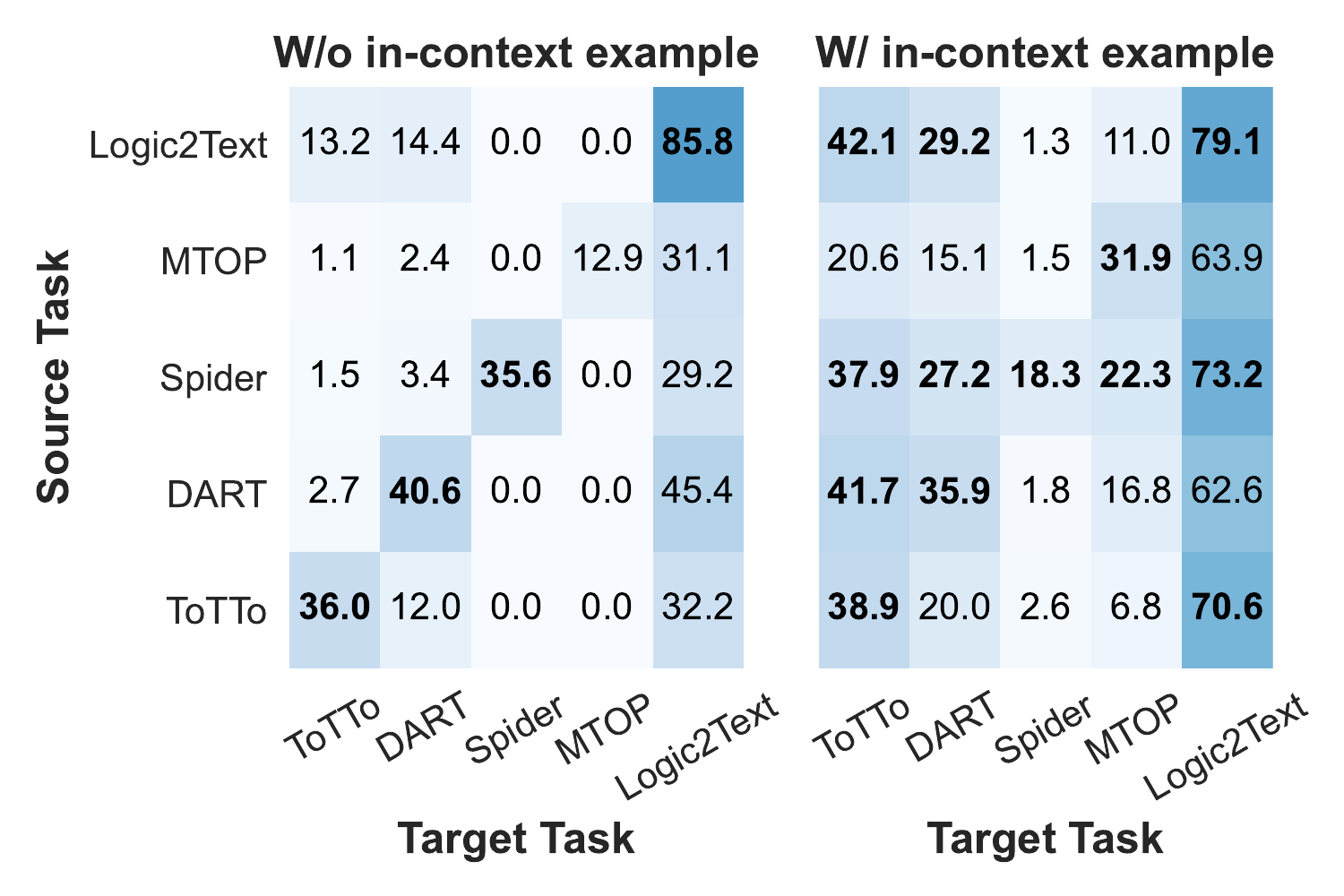}
    \caption{Cross-task evaluation of prompt tuning with (right) and without (left) a target in-context example.. Numbers better than the corresponding ICL baseline for the target task are bolded. Pairing source task embeddings with target task in-context demonstrations increases task transfer.}
    \label{fig:cross_task}
\end{figure}

\paragraph{IPT is more stable than PT with more soft prompt tokens:} \label{sec:analysis_var}
The number of soft prompt tokens in PT is an important hyperparameters: while more prompt tokens may endow the model with increased capacity to adapt to downstream tasks, it also becomes harder to optimize. As shown in Figure~\ref{fig:vpt}, average PT performance first increases and then decreases as the number of prompt tokens increases from 1 to 150. However, the variance of PT consistently increases as the number of prompt tokens increases\footnote{The high variance of PT is also observed in prior works~\citep{channel-pt,tu-spot,vu-etal-2022-overcoming}}. On the other hand, the discrete in-context example in IPT improves the method's stability with more prompt tokens, and IPT also reaches its best performance with more soft prompt tokens than PT. We conjecture that additional parameters (i.e., soft prompt tokens) are necessary to learn how to integrate the dynamically-retrieved in-context demonstrations. Overall, IPT's improved stability is a clear positive especially when applying parameter-efficient tuning methods to large LMs, where hyperparameter selection can be computationally infeasible. 

\paragraph{Prompt embeddings are transferable to new tasks provided with in-context demonstrations}

Adaptation methods that require training, such as PT or IPT, still require a large amount of labeled data for the target task, which is not available in low-resource settings. 
Thus, we measure how much soft prompts learned for a \emph{source} task can help improve performance on a different \emph{target} task for which it may not be possible to learn a powerful soft prompt. We simulate this setting by conducting cross-task evaluations\footnote{We present another analysis on the input format transferability in Appendix~\ref{sec:input_transfer}.} across our five tasks, measuring whether soft prompts learned by PT can transfer to other tasks when paired with an in-context demonstration from the target task. Figure~\ref{fig:cross_task} shows that embeddings learned via PT alone are generally not transferable to new tasks, as evidenced by the low off-diagonal numbers in the left heatmap (bolded values represent better performance than ICL). However, pairing the prompt embeddings learned on a source task with a target task in-context demonstration often performs better than just the latter (right heatmap). For instance, while vanilla ICL on ToTTo achieves 35.1 BLEU, pairing the in-context demonstration with a soft prompt learned on DART boosts performance to 41.7. 
These results show that although the learned prompt embeddings are task-specific, they encode information applicable to other tasks and help take better advantage of in-context demonstrations. 

\section{Conclusion}
In this paper, we empirically analyze the effect of in-context demonstrations on prompt tuning for five language generation tasks. Our experiments reveal that while instruction prompt tuning and prompt tuning perform competitively with each other, IPT is more stable, yielding lower variance when varying hyperparameters. IPT also significantly improves over PT when the in-context demonstration closely resembles the test input, which is frequently the case in the ToTTo dataset. Finally, soft prompts learned for a source task can exhibit positive transfer to new target tasks when paired with in-context demonstrations.

\section*{Limitation}

While we have examined the interplay of prompt tuning and in-context learning on more general datasets than previous work, our experiments were limited to only $~$1B parameter language models. Future research on larger models is necessary to see if our findings scale with parameter count. Although we haven shown instruction prompt tuning is more stable than prompt tuning, its training is also slower than vanilla prompt tuning.


\bibliography{anthology,custom}

\begin{thebibliography}{33}
\expandafter\ifx\csname natexlab\endcsname\relax\def\natexlab#1{#1}\fi

\bibitem[{An et~al.(2022)An, Li, Lin, Liu, Chen, Fu, Chen, Zheng, and
  Lou}]{input-tuning}
Shengnan An, Yifei Li, Zeqi Lin, Qian Liu, Bei Chen, Qiang Fu, Weizhu Chen,
  Nanning Zheng, and Jian-Guang Lou. 2022.
\newblock \href {https://doi.org/10.48550/ARXIV.2203.03131} {Input-tuning:
  Adapting unfamiliar inputs to frozen pretrained models}.

\bibitem[{Bapna and Firat(2019)}]{adapter-nmt}
Ankur Bapna and Orhan Firat. 2019.
\newblock \href {https://doi.org/10.18653/v1/D19-1165} {Simple, scalable
  adaptation for neural machine translation}.
\newblock In \emph{Proceedings of the 2019 Conference on Empirical Methods in
  Natural Language Processing and the 9th International Joint Conference on
  Natural Language Processing (EMNLP-IJCNLP)}, pages 1538--1548, Hong Kong,
  China. Association for Computational Linguistics.

\bibitem[{Ben~Zaken et~al.(2022)Ben~Zaken, Goldberg, and Ravfogel}]{bitfit}
Elad Ben~Zaken, Yoav Goldberg, and Shauli Ravfogel. 2022.
\newblock \href {https://doi.org/10.18653/v1/2022.acl-short.1} {{B}it{F}it:
  Simple parameter-efficient fine-tuning for transformer-based masked
  language-models}.
\newblock In \emph{Proceedings of the 60th Annual Meeting of the Association
  for Computational Linguistics (Volume 2: Short Papers)}, pages 1--9, Dublin,
  Ireland. Association for Computational Linguistics.

\bibitem[{Brown et~al.(2020)Brown, Mann, Ryder, Subbiah, Kaplan, Dhariwal,
  Neelakantan, Shyam, Sastry, Askell, Agarwal, Herbert-Voss, Krueger, Henighan,
  Child, Ramesh, Ziegler, Wu, Winter, Hesse, Chen, Sigler, Litwin, Gray, Chess,
  Clark, Berner, McCandlish, Radford, Sutskever, and Amodei}]{gpt3}
Tom~B. Brown, Benjamin Mann, Nick Ryder, Melanie Subbiah, Jared Kaplan,
  Prafulla Dhariwal, Arvind Neelakantan, Pranav Shyam, Girish Sastry, Amanda
  Askell, Sandhini Agarwal, Ariel Herbert-Voss, Gretchen Krueger, Tom Henighan,
  Rewon Child, Aditya Ramesh, Daniel~M. Ziegler, Jeffrey Wu, Clemens Winter,
  Christopher Hesse, Mark Chen, Eric Sigler, Mateusz Litwin, Scott Gray,
  Benjamin Chess, Jack Clark, Christopher Berner, Sam McCandlish, Alec Radford,
  Ilya Sutskever, and Dario Amodei. 2020.
\newblock \href {https://doi.org/10.48550/ARXIV.2005.14165} {Language models
  are few-shot learners}.

\bibitem[{Chen et~al.(2022)Chen, Zhong, Zha, Karypis, and
  He}]{chen-etal-2022-meta}
Yanda Chen, Ruiqi Zhong, Sheng Zha, George Karypis, and He~He. 2022.
\newblock \href {https://doi.org/10.18653/v1/2022.acl-long.53} {Meta-learning
  via language model in-context tuning}.
\newblock In \emph{Proceedings of the 60th Annual Meeting of the Association
  for Computational Linguistics (Volume 1: Long Papers)}, pages 719--730,
  Dublin, Ireland. Association for Computational Linguistics.

\bibitem[{Chen et~al.(2020)Chen, Chen, Zha, Zhou, Zhang, Sundaresan, and
  Wang}]{logic2text}
Zhiyu Chen, Wenhu Chen, Hanwen Zha, Xiyou Zhou, Yunkai Zhang, Sairam
  Sundaresan, and William~Yang Wang. 2020.
\newblock \href {https://doi.org/10.18653/v1/2020.findings-emnlp.190}
  {{L}ogic2{T}ext: High-fidelity natural language generation from logical
  forms}.
\newblock In \emph{Findings of the Association for Computational Linguistics:
  EMNLP 2020}, pages 2096--2111, Online. Association for Computational
  Linguistics.

\bibitem[{Chowdhery et~al.(2022)Chowdhery, Narang, Devlin, Bosma, Mishra,
  Roberts, Barham, Chung, Sutton, Gehrmann et~al.}]{chowdhery2022palm}
Aakanksha Chowdhery, Sharan Narang, Jacob Devlin, Maarten Bosma, Gaurav Mishra,
  Adam Roberts, Paul Barham, Hyung~Won Chung, Charles Sutton, Sebastian
  Gehrmann, et~al. 2022.
\newblock Palm: Scaling language modeling with pathways.
\newblock \emph{arXiv preprint arXiv:2204.02311}.

\bibitem[{Ding et~al.(2022)Ding, Qin, Yang, Wei, Yang, Su, Hu, Chen, Chan,
  Chen, Yi, Zhao, Wang, Liu, Zheng, Chen, Liu, Tang, Li, and Sun}]{peft-study}
Ning Ding, Yujia Qin, Guang Yang, Fuchao Wei, Zonghan Yang, Yusheng Su,
  Shengding Hu, Yulin Chen, Chi-Min Chan, Weize Chen, Jing Yi, Weilin Zhao,
  Xiaozhi Wang, Zhiyuan Liu, Hai-Tao Zheng, Jianfei Chen, Yang Liu, Jie Tang,
  Juanzi Li, and Maosong Sun. 2022.
\newblock \href {https://doi.org/10.48550/ARXIV.2203.06904} {Delta tuning: A
  comprehensive study of parameter efficient methods for pre-trained language
  models}.

\bibitem[{Gu et~al.(2022)Gu, Han, Liu, and Huang}]{ppt}
Yuxian Gu, Xu~Han, Zhiyuan Liu, and Minlie Huang. 2022.
\newblock \href {https://doi.org/10.18653/v1/2022.acl-long.576} {{PPT}:
  Pre-trained prompt tuning for few-shot learning}.
\newblock In \emph{Proceedings of the 60th Annual Meeting of the Association
  for Computational Linguistics (Volume 1: Long Papers)}, pages 8410--8423,
  Dublin, Ireland. Association for Computational Linguistics.

\bibitem[{Han et~al.(2021)Han, Zhao, Ding, Liu, and Sun}]{ptr}
Xu~Han, Weilin Zhao, Ning Ding, Zhiyuan Liu, and Maosong Sun. 2021.
\newblock \href {https://doi.org/10.48550/ARXIV.2105.11259} {Ptr: Prompt tuning
  with rules for text classification}.

\bibitem[{Houlsby et~al.(2019)Houlsby, Giurgiu, Jastrzebski, Morrone,
  De~Laroussilhe, Gesmundo, Attariyan, and Gelly}]{pmlr-v97-houlsby19a}
Neil Houlsby, Andrei Giurgiu, Stanislaw Jastrzebski, Bruna Morrone, Quentin
  De~Laroussilhe, Andrea Gesmundo, Mona Attariyan, and Sylvain Gelly. 2019.
\newblock \href {https://proceedings.mlr.press/v97/houlsby19a.html}
  {Parameter-efficient transfer learning for {NLP}}.
\newblock In \emph{Proceedings of the 36th International Conference on Machine
  Learning}, volume~97 of \emph{Proceedings of Machine Learning Research},
  pages 2790--2799. PMLR.

\bibitem[{Johnson et~al.(2019)Johnson, Douze, and
  J{\'e}gou}]{johnson2019billion}
Jeff Johnson, Matthijs Douze, and Herv{\'e} J{\'e}gou. 2019.
\newblock Billion-scale similarity search with {GPUs}.
\newblock \emph{IEEE Transactions on Big Data}, 7(3):535--547.

\bibitem[{Karimi~Mahabadi et~al.(2021)Karimi~Mahabadi, Ruder, Dehghani, and
  Henderson}]{karimi-mahabadi-etal-2021-parameter}
Rabeeh Karimi~Mahabadi, Sebastian Ruder, Mostafa Dehghani, and James Henderson.
  2021.
\newblock \href {https://doi.org/10.18653/v1/2021.acl-long.47}
  {Parameter-efficient multi-task fine-tuning for transformers via shared
  hypernetworks}.
\newblock In \emph{Proceedings of the 59th Annual Meeting of the Association
  for Computational Linguistics and the 11th International Joint Conference on
  Natural Language Processing (Volume 1: Long Papers)}, pages 565--576, Online.
  Association for Computational Linguistics.

\bibitem[{Lester et~al.(2021)Lester, Al-Rfou, and Constant}]{lester-pt}
Brian Lester, Rami Al-Rfou, and Noah Constant. 2021.
\newblock \href {https://doi.org/10.18653/v1/2021.emnlp-main.243} {The power of
  scale for parameter-efficient prompt tuning}.
\newblock In \emph{Proceedings of the 2021 Conference on Empirical Methods in
  Natural Language Processing}, pages 3045--3059, Online and Punta Cana,
  Dominican Republic. Association for Computational Linguistics.

\bibitem[{Li et~al.(2021)Li, Arora, Chen, Gupta, Gupta, and Mehdad}]{mtop}
Haoran Li, Abhinav Arora, Shuohui Chen, Anchit Gupta, Sonal Gupta, and Yashar
  Mehdad. 2021.
\newblock \href {https://doi.org/10.18653/v1/2021.eacl-main.257} {{MTOP}: A
  comprehensive multilingual task-oriented semantic parsing benchmark}.
\newblock In \emph{Proceedings of the 16th Conference of the European Chapter
  of the Association for Computational Linguistics: Main Volume}, pages
  2950--2962, Online. Association for Computational Linguistics.

\bibitem[{Li and Liang(2021)}]{prefix-tune}
Xiang~Lisa Li and Percy Liang. 2021.
\newblock \href {https://doi.org/10.18653/v1/2021.acl-long.353} {Prefix-tuning:
  Optimizing continuous prompts for generation}.
\newblock In \emph{Proceedings of the 59th Annual Meeting of the Association
  for Computational Linguistics and the 11th International Joint Conference on
  Natural Language Processing (Volume 1: Long Papers)}, pages 4582--4597,
  Online. Association for Computational Linguistics.

\bibitem[{Liu et~al.(2022{\natexlab{a}})Liu, Tam, Mohammed, Mohta, Huang,
  Bansal, and Raffel}]{liu2022fewshot}
Haokun Liu, Derek Tam, Muqeeth Mohammed, Jay Mohta, Tenghao Huang, Mohit
  Bansal, and Colin Raffel. 2022{\natexlab{a}}.
\newblock \href {https://openreview.net/forum?id=rBCvMG-JsPd} {Few-shot
  parameter-efficient fine-tuning is better and cheaper than in-context
  learning}.
\newblock In \emph{Advances in Neural Information Processing Systems}.

\bibitem[{Liu et~al.(2022{\natexlab{b}})Liu, Shen, Zhang, Dolan, Carin, and
  Chen}]{good-gpt3-example}
Jiachang Liu, Dinghan Shen, Yizhe Zhang, Bill Dolan, Lawrence Carin, and Weizhu
  Chen. 2022{\natexlab{b}}.
\newblock \href {https://doi.org/10.18653/v1/2022.deelio-1.10} {What makes good
  in-context examples for {GPT}-3?}
\newblock In \emph{Proceedings of Deep Learning Inside Out (DeeLIO 2022): The
  3rd Workshop on Knowledge Extraction and Integration for Deep Learning
  Architectures}, pages 100--114, Dublin, Ireland and Online. Association for
  Computational Linguistics.

\bibitem[{Liu et~al.(2022{\natexlab{c}})Liu, Ji, Fu, Tam, Du, Yang, and
  Tang}]{p-tuning-v2}
Xiao Liu, Kaixuan Ji, Yicheng Fu, Weng Tam, Zhengxiao Du, Zhilin Yang, and Jie
  Tang. 2022{\natexlab{c}}.
\newblock \href {https://doi.org/10.18653/v1/2022.acl-short.8} {{P}-tuning:
  Prompt tuning can be comparable to fine-tuning across scales and tasks}.
\newblock In \emph{Proceedings of the 60th Annual Meeting of the Association
  for Computational Linguistics (Volume 2: Short Papers)}, pages 61--68,
  Dublin, Ireland. Association for Computational Linguistics.

\bibitem[{Liu et~al.(2021)Liu, Zheng, Du, Ding, Qian, Yang, and
  Tang}]{p-tuning}
Xiao Liu, Yanan Zheng, Zhengxiao Du, Ming Ding, Yujie Qian, Zhilin Yang, and
  Jie Tang. 2021.
\newblock \href {https://doi.org/10.48550/ARXIV.2103.10385} {Gpt understands,
  too}.

\bibitem[{Loshchilov and Hutter(2019)}]{adamw}
Ilya Loshchilov and Frank Hutter. 2019.
\newblock \href {https://openreview.net/forum?id=Bkg6RiCqY7} {Decoupled weight
  decay regularization}.
\newblock In \emph{International Conference on Learning Representations}.

\bibitem[{Min et~al.(2022{\natexlab{a}})Min, Lewis, Hajishirzi, and
  Zettlemoyer}]{channel-pt}
Sewon Min, Mike Lewis, Hannaneh Hajishirzi, and Luke Zettlemoyer.
  2022{\natexlab{a}}.
\newblock \href {https://doi.org/10.18653/v1/2022.acl-long.365} {Noisy channel
  language model prompting for few-shot text classification}.
\newblock In \emph{Proceedings of the 60th Annual Meeting of the Association
  for Computational Linguistics (Volume 1: Long Papers)}, pages 5316--5330,
  Dublin, Ireland. Association for Computational Linguistics.

\bibitem[{Min et~al.(2022{\natexlab{b}})Min, Lewis, Zettlemoyer, and
  Hajishirzi}]{min-etal-2022-metaicl}
Sewon Min, Mike Lewis, Luke Zettlemoyer, and Hannaneh Hajishirzi.
  2022{\natexlab{b}}.
\newblock \href {https://doi.org/10.18653/v1/2022.naacl-main.201} {{M}eta{ICL}:
  Learning to learn in context}.
\newblock In \emph{Proceedings of the 2022 Conference of the North American
  Chapter of the Association for Computational Linguistics: Human Language
  Technologies}, pages 2791--2809, Seattle, United States. Association for
  Computational Linguistics.

\bibitem[{Nan et~al.(2021)Nan, Radev, Zhang, Rau, Sivaprasad, Hsieh, Tang,
  Vyas, Verma, Krishna, Liu, Irwanto, Pan, Rahman, Zaidi, Mutuma, Tarabar,
  Gupta, Yu, Tan, Lin, Xiong, Socher, and Rajani}]{dart}
Linyong Nan, Dragomir Radev, Rui Zhang, Amrit Rau, Abhinand Sivaprasad,
  Chiachun Hsieh, Xiangru Tang, Aadit Vyas, Neha Verma, Pranav Krishna,
  Yangxiaokang Liu, Nadia Irwanto, Jessica Pan, Faiaz Rahman, Ahmad Zaidi,
  Mutethia Mutuma, Yasin Tarabar, Ankit Gupta, Tao Yu, Yi~Chern Tan,
  Xi~Victoria Lin, Caiming Xiong, Richard Socher, and Nazneen~Fatema Rajani.
  2021.
\newblock \href {https://doi.org/10.18653/v1/2021.naacl-main.37} {{DART}:
  Open-domain structured data record to text generation}.
\newblock In \emph{Proceedings of the 2021 Conference of the North American
  Chapter of the Association for Computational Linguistics: Human Language
  Technologies}, pages 432--447, Online. Association for Computational
  Linguistics.

\bibitem[{Papineni et~al.(2002)Papineni, Roukos, Ward, and Zhu}]{bleu}
Kishore Papineni, Salim Roukos, Todd Ward, and Wei-Jing Zhu. 2002.
\newblock \href {https://doi.org/10.3115/1073083.1073135} {{B}leu: a method for
  automatic evaluation of machine translation}.
\newblock In \emph{Proceedings of the 40th Annual Meeting of the Association
  for Computational Linguistics}, pages 311--318, Philadelphia, Pennsylvania,
  USA. Association for Computational Linguistics.

\bibitem[{Parikh et~al.(2020)Parikh, Wang, Gehrmann, Faruqui, Dhingra, Yang,
  and Das}]{totto}
Ankur Parikh, Xuezhi Wang, Sebastian Gehrmann, Manaal Faruqui, Bhuwan Dhingra,
  Diyi Yang, and Dipanjan Das. 2020.
\newblock \href {https://doi.org/10.18653/v1/2020.emnlp-main.89} {{ToTTo}: A
  controlled table-to-text generation dataset}.
\newblock In \emph{Proceedings of the 2020 Conference on Empirical Methods in
  Natural Language Processing (EMNLP)}, pages 1173--1186, Online. Association
  for Computational Linguistics.

\bibitem[{Singhal et~al.(2022)Singhal, Azizi, Tu, Mahdavi, Wei, Chung, Scales,
  Tanwani, Cole-Lewis, Pfohl, Payne, Seneviratne, Gamble, Kelly, Scharli,
  Chowdhery, Mansfield, Arcas, Webster, Corrado, Matias, Chou, Gottweis,
  Tomasev, Liu, Rajkomar, Barral, Semturs, Karthikesalingam, and
  Natarajan}]{llm-clinical}
Karan Singhal, Shekoofeh Azizi, Tao Tu, S.~Sara Mahdavi, Jason Wei, Hyung~Won
  Chung, Nathan Scales, Ajay Tanwani, Heather Cole-Lewis, Stephen Pfohl, Perry
  Payne, Martin Seneviratne, Paul Gamble, Chris Kelly, Nathaneal Scharli,
  Aakanksha Chowdhery, Philip Mansfield, Blaise Aguera~y Arcas, Dale Webster,
  Greg~S. Corrado, Yossi Matias, Katherine Chou, Juraj Gottweis, Nenad Tomasev,
  Yun Liu, Alvin Rajkomar, Joelle Barral, Christopher Semturs, Alan
  Karthikesalingam, and Vivek Natarajan. 2022.
\newblock \href {https://doi.org/10.48550/ARXIV.2212.13138} {Large language
  models encode clinical knowledge}.

\bibitem[{Vu et~al.(2022{\natexlab{a}})Vu, Barua, Lester, Cer, Iyyer, and
  Constant}]{vu-etal-2022-overcoming}
Tu~Vu, Aditya Barua, Brian Lester, Daniel Cer, Mohit Iyyer, and Noah Constant.
  2022{\natexlab{a}}.
\newblock \href {https://aclanthology.org/2022.emnlp-main.630} {Overcoming
  catastrophic forgetting in zero-shot cross-lingual generation}.
\newblock In \emph{Proceedings of the 2022 Conference on Empirical Methods in
  Natural Language Processing}, pages 9279--9300, Abu Dhabi, United Arab
  Emirates. Association for Computational Linguistics.

\bibitem[{Vu et~al.(2022{\natexlab{b}})Vu, Lester, Constant, Al-Rfou{'}, and
  Cer}]{tu-spot}
Tu~Vu, Brian Lester, Noah Constant, Rami Al-Rfou{'}, and Daniel Cer.
  2022{\natexlab{b}}.
\newblock \href {https://doi.org/10.18653/v1/2022.acl-long.346} {{SP}o{T}:
  Better frozen model adaptation through soft prompt transfer}.
\newblock In \emph{Proceedings of the 60th Annual Meeting of the Association
  for Computational Linguistics (Volume 1: Long Papers)}, pages 5039--5059,
  Dublin, Ireland. Association for Computational Linguistics.

\bibitem[{Wei et~al.(2022)Wei, Wang, Schuurmans, Bosma, brian ichter, Xia, Chi,
  Le, and Zhou}]{wei2022chain}
Jason Wei, Xuezhi Wang, Dale Schuurmans, Maarten Bosma, brian ichter, Fei Xia,
  Ed~H. Chi, Quoc~V Le, and Denny Zhou. 2022.
\newblock \href {https://openreview.net/forum?id=_VjQlMeSB_J} {Chain of thought
  prompting elicits reasoning in large language models}.
\newblock In \emph{Advances in Neural Information Processing Systems}.

\bibitem[{Xie et~al.(2022)Xie, Wu, Shi, Zhong, Scholak, Yasunaga, Wu, Zhong,
  Yin, Wang, Zhong, Wang, Li, Boyle, Ni, Yao, Radev, Xiong, Kong, Zhang, Smith,
  Zettlemoyer, and Yu}]{unifiedskg}
Tianbao Xie, Chen~Henry Wu, Peng Shi, Ruiqi Zhong, Torsten Scholak, Michihiro
  Yasunaga, Chien-Sheng Wu, Ming Zhong, Pengcheng Yin, Sida~I. Wang, Victor
  Zhong, Bailin Wang, Chengzu Li, Connor Boyle, Ansong Ni, Ziyu Yao, Dragomir
  Radev, Caiming Xiong, Lingpeng Kong, Rui Zhang, Noah~A. Smith, Luke
  Zettlemoyer, and Tao Yu. 2022.
\newblock \href {https://aclanthology.org/2022.emnlp-main.39} {{U}nified{SKG}:
  Unifying and multi-tasking structured knowledge grounding with text-to-text
  language models}.
\newblock In \emph{Proceedings of the 2022 Conference on Empirical Methods in
  Natural Language Processing}, pages 602--631, Abu Dhabi, United Arab
  Emirates. Association for Computational Linguistics.

\bibitem[{Yu et~al.(2018)Yu, Zhang, Yang, Yasunaga, Wang, Li, Ma, Li, Yao,
  Roman, Zhang, and Radev}]{spider}
Tao Yu, Rui Zhang, Kai Yang, Michihiro Yasunaga, Dongxu Wang, Zifan Li, James
  Ma, Irene Li, Qingning Yao, Shanelle Roman, Zilin Zhang, and Dragomir Radev.
  2018.
\newblock \href {https://doi.org/10.18653/v1/D18-1425} {{S}pider: A large-scale
  human-labeled dataset for complex and cross-domain semantic parsing and
  text-to-{SQL} task}.
\newblock In \emph{Proceedings of the 2018 Conference on Empirical Methods in
  Natural Language Processing}, pages 3911--3921, Brussels, Belgium.
  Association for Computational Linguistics.

\bibitem[{Zhou et~al.(2022)Zhou, Schärli, Hou, Wei, Scales, Wang, Schuurmans,
  Cui, Bousquet, Le, and Chi}]{least-to-most}
Denny Zhou, Nathanael Schärli, Le~Hou, Jason Wei, Nathan Scales, Xuezhi Wang,
  Dale Schuurmans, Claire Cui, Olivier Bousquet, Quoc Le, and Ed~Chi. 2022.
\newblock \href {https://doi.org/10.48550/ARXIV.2205.10625} {Least-to-most
  prompting enables complex reasoning in large language models}.

\end{thebibliography}
\bibliographystyle{acl_natbib}

\appendix
\begin{table*}[]
    \centering
    \scalebox{0.85}{\begin{tabular}{@{}ll@{}}
\toprule
\textbf{Task} & \textbf{Input format}                       \\ \midrule
ToTTo         & \textcolor{blue}{Table}:{[}linearized table{]}\textcolor{blue}{Sentence}:{[}output{]}\textcolor{gray}{\textbackslash{}n}\textcolor{gray}{\textbackslash{}n}\textcolor{blue}{Table}:{[}linearized table{]}\textcolor{blue}{Sentence}:                          \\
DART          & \textcolor{blue}{Table}:{[}linearized table{]}\textcolor{blue}{Text}:{[}output{]}\textcolor{gray}{\textbackslash{}n}\textcolor{gray}{\textbackslash{}n}\textcolor{blue}{Table}:{[}linearized table{]}\textcolor{blue}{Text}: \\
Spider        & \textcolor{blue}{Input}:{[}table schema{]}\textcolor{gray}{\textbackslash{}t}{[}input string{]}\textcolor{blue}{Output}:{[}SQL{]}\textcolor{gray}{\textbackslash{}n}\textcolor{gray}{\textbackslash{}n}\textcolor{blue}{Input}:{[}table schema{]}\textcolor{gray}{\textbackslash{}t}{[}input string{]}\textcolor{blue}{Output}:    \\
MTOP          & \textcolor{blue}{Input}:{[}API calls{]}\textcolor{gray}{\textbackslash{}t}{[}input string{]}\textcolor{blue}{Output}:{[}output{]}\textcolor{gray}{\textbackslash{}n}\textcolor{gray}{\textbackslash{}n}\textcolor{blue}{Input}:{[}API calls{]}\textcolor{gray}{\textbackslash{}t}{[}input string{]}\textcolor{blue}{Output}:       \\
Logic2Text    & \textcolor{blue}{Input}:{[}table schema{]}\textcolor{gray}{\textbackslash{}t}{[}input string{]}\textcolor{blue}{Output}:{[}output{]}\textcolor{gray}{\textbackslash{}n}\textcolor{gray}{\textbackslash{}n}\textcolor{blue}{Input}:{[}table schema{]}\textcolor{gray}{\textbackslash{}t}{[}input string{]}\textcolor{blue}{Output}: \\ \bottomrule
\end{tabular}}
    \caption{The input format of each task for instruction prompt tuning and in-context learning. Soft prompts for IPT is ommited in the table.}
    \label{tab:ipt-format}
\end{table*}
\section{Model Hyperparameters} \label{sec:hparam}

For both prompt tuning and instruction prompt tuning, we set batch size 8 and grid search learning rate over $\{5e-5, 5e-4, 1e-3\}$ and weight decay over $\{0.0, 0.01, 0.1\}$. The adopted hyperparameters for each task and each approach is presented in Table~\ref{tab:hparam}.

\begin{table}[!h]
    \centering
    \footnotesize
    \begin{tabular}{cccccc}
\hline
\multicolumn{1}{l}{}   & \multicolumn{1}{l}{} & \multicolumn{2}{c}{PT} & \multicolumn{2}{c}{IPT} \\ \hline
\multicolumn{1}{l}{}   & Task                 & lr        & decay      & lr         & decay      \\
\multirow{5}{*}{BLOOM} & ToTTo                & 5e-5      & 0.0        & 5e-5       & 0.01       \\
                       & Dart                 & 5e-5      & 0.0        & 5e-5       & 0.0        \\
                       & Spider               & 5e-5      & 0.1        & 5e-5       & 0.1        \\
                       & MTOP                 & 5e-4      & 0.0        & 5e-4       & 0.01       \\
                       & Logic2Text           & 5e-4      & 0.01       & 5e-4       & 0.0        \\ \hline
\multirow{5}{*}{OPT}   & ToTTo                & 5e-5      & 0.0        & 5e-5       & 0.0        \\
                       & Dart                 & 5e-5      & 0.0        & 5e-5       & 0.0        \\
                       & Spider               & 5e-4      & 0.0        & 5e-4       & 0.0        \\
                       & MTOP                 & 5e-4      & 0.01       & 5e-5       & 0.0        \\
                       & Logic2Text           & 5e-4      & 0.0        & 5e-4       & 0.0        \\ \hline
\multirow{5}{*}{GPT2}  & ToTTo                & 5e-5      & 0.0        & 5e-5       & 0.0        \\
                       & Dart                 & 5e-5      & 0.0        & 5e-5       & 0.0        \\
                       & Spider               & 5e-5      & 0.0        & 5e-5       & 0.0        \\
                       & MTOP                 & 5e-4      & 0.01        & 5e-4       & 0.01        \\
                       & Logic2Text           & 5e-4      & 0.0        & 5e-4       & 0.0        \\ \hline
\end{tabular}
    \caption{Hyperparameters of PT and IPT for each task.}
    \label{tab:hparam}
\end{table}

The input format of IPT for each task is presented in Table~\ref{tab:ipt-format}. 

\section{Retrieve in-context demonstration for DART} \label{sec:appendix-dart}
As DART contains examples sharing the same input, i.e., the same input corresponds to different outputs, examples having the same input will be selected as the in-context demonstration of each other. However, our earlier experiments indicated that prepending these examples leads to convergence to higher losses, and worse performance overall on evaluation set. Therefore, for this dataset, we exclude same-input examples and select the top semantically-similar examples from the rest as in-context demonstration.

\section{Cross Input Transferability} \label{sec:input_transfer}
\begin{table}[!t]
\scalebox{0.9}{
\begin{tabular}{@{}lcccc@{}}
\toprule
           & PT   & IPT w/o ICL & PT w/ ICL & IPT  \\ \midrule
ToTTo      & \underline{36.0} & 16.7        & \underline{38.9}      & \underline{47.1} \\
DART       & \underline{40.6} & 4.4         & \underline{35.9}      & \underline{41.3} \\

Spider     & \underline{35.6} & \underline{24.9}        & \underline{18.3}      & \underline{33.6} \\
MTOP       & 12.9 & 0.0         & \underline{31.9}      & \underline{60.5} \\ 
Logic2Text & \underline{85.8} & \underline{72.9}        & \underline{79.1}      & \underline{85.8} \\\bottomrule
\end{tabular}
}
\caption{Instruction prompt tuning performs worse when the in-context demonstration is removed (second column), whereas regular PT embeddings adapt better to input with in-context demonstration (third column). Performance better than retrieved one-shot ICL is underlined.}
\label{tab:input_transfer}
\end{table}
We explore how well the embeddings learned via PT and IPT can adapt to the input setting of each other. Table~\ref{tab:input_transfer} shows that the performance of IPT drops significantly when the in-context demonstrations are removed, indicating the critical role of these demonstrations in IPT. Appending regular PT embeddings with in-context demonstration leads to on average smaller degradation in performance, and outperforms one-shot ICL consistently across all tasks. On two datasets (ToTTo and MTOP), having in-context demonstrations exceeds the performance of regular prompt tuning, suggesting that retrieved demonstrations can provide necessary signals that are not well-encoded into the soft prompts. 

\end{document}